\documentclass[conference]{IEEEtran}
\IEEEoverridecommandlockouts
\usepackage{cite}
\usepackage{amsmath,amssymb,amsfonts}
\allowdisplaybreaks
\usepackage{algorithmic}
\usepackage{graphicx}
\usepackage{textcomp}
\usepackage{booktabs}
\usepackage{colortbl}
\usepackage[table,xcdraw]{xcolor}
\usepackage{multirow}
\usepackage{marvosym}
\def\BibTeX{{\rm B\kern-.05em{\sc i\kern-.025em b}\kern-.08em
    T\kern-.1667em\lower.7ex\hbox{E}\kern-.125emX}}

\begin{document}

\title{MixPolyp: Integrating Mask, Box and Scribble Supervision for Enhanced Polyp Segmentation}

\author{
    \IEEEauthorblockN{1\textsuperscript{st} Yiwen Hu$\dagger$\thanks{$\dagger$~Equal contributions.}}
    \IEEEauthorblockA{
        \textit{FNii, CUHK-Shenzhen} \\
        \textit{SSE, CUHK-Shenzhen}\\
        \textit{South China Hospital, Shenzhen University}\\       
        Shenzhen, China \\
        yiwenhu1@link.cuhk.edu.cn
    }
    \and
    \IEEEauthorblockN{2\textsuperscript{nd} Jun Wei$\dagger$}
    \IEEEauthorblockA{
        \textit{FNii, CUHK-Shenzhen} \\
        \textit{SSE, CUHK-Shenzhen}\\
        Shenzhen, China \\
        junwei@link.cuhk.edu.cn
    }    
    \and
    \IEEEauthorblockN{3\textsuperscript{rd} Yuncheng Jiang$\dagger$}
    \IEEEauthorblockA{
        \textit{FNii, CUHK-Shenzhen} \\
        \textit{SSE, CUHK-Shenzhen}\\
        \textit{SRIBD, Shenzhen}\\
        Shenzhen, China \\
        yunchengjiang@link.cuhk.edu.cn
    }
    \and
    \IEEEauthorblockN{4\textsuperscript{th} Haoyang Li, }
    \IEEEauthorblockA{
        \textit{FNii, CUHK-Shenzhen} \\
        \textit{SDS, CUHK-Shenzhen}\\
        Shenzhen, China \\
        haoyangli@link.cuhk.edu.cn
    }    
    \and
    \IEEEauthorblockN{5\textsuperscript{th} Shuguang Cui}
    \IEEEauthorblockA{
        \textit{SSE, CUHK-Shenzhen}\\
        \textit{FNii, CUHK-Shenzhen} \\
        Shenzhen, China \\
        shuguangcui@cuhk.edu.cn
    }
    \and    
    \IEEEauthorblockN{6\textsuperscript{th} Zhen Li\textsuperscript{\Letter}\thanks{\textsuperscript{\Letter}~Corresponding authors}}
    \IEEEauthorblockA{
        \textit{SSE, CUHK-Shenzhen}\\
        \textit{FNii, CUHK-Shenzhen} \\
        Shenzhen, China \\
        lizhen@cuhk.edu.cn
    }
    \and
    \IEEEauthorblockN{7\textsuperscript{th} Song Wu\textsuperscript{\Letter}}
    \IEEEauthorblockA{
        \textit{South China Hospital} \\
        \textit{Health Science Center}\\
        \textit{Shenzhen University} \\
        Shenzhen, China \\
        wusong@szu.edu.cn
    }
}

\maketitle

\begin{abstract}
Limited by the expensive labeling, polyp segmentation models are plagued by data shortages. To tackle this, we propose the mixed supervised polyp segmentation paradigm (MixPolyp). Unlike traditional models relying on a single type of annotation, MixPolyp combines diverse annotation types (mask, box, and scribble) within a single model, thereby expanding the range of available data and reducing labeling costs. To achieve this, MixPolyp introduces three novel supervision losses to handle various annotations: Subspace Projection loss ($\mathcal{L_{SP}}$), Binary Minimum Entropy loss ($\mathcal{L_{BME}}$), and Linear Regularization loss ($\mathcal{L_{LR}}$). For box annotations, $\mathcal{L_{SP}}$ eliminates shape inconsistencies between the prediction and the supervision. For scribble annotations, $\mathcal{L_{BME}}$ provides supervision for unlabeled pixels through minimum entropy constraint, thereby alleviating supervision sparsity. Furthermore, $\mathcal{L_{LR}}$ provides dense supervision by enforcing consistency among the predictions, thus reducing the non-uniqueness. These losses are independent of the model structure, making them generally applicable. They are used only during training, adding no computational cost during inference. Extensive experiments on five datasets demonstrate MixPolyp's effectiveness.
\end{abstract}

\begin{IEEEkeywords}
Polyp Segmentation, Mixed Supervision, Efficient Annotation
\end{IEEEkeywords}

\section{Introduction}

\begin{figure*}[t]
\centering
\includegraphics[width=\linewidth]{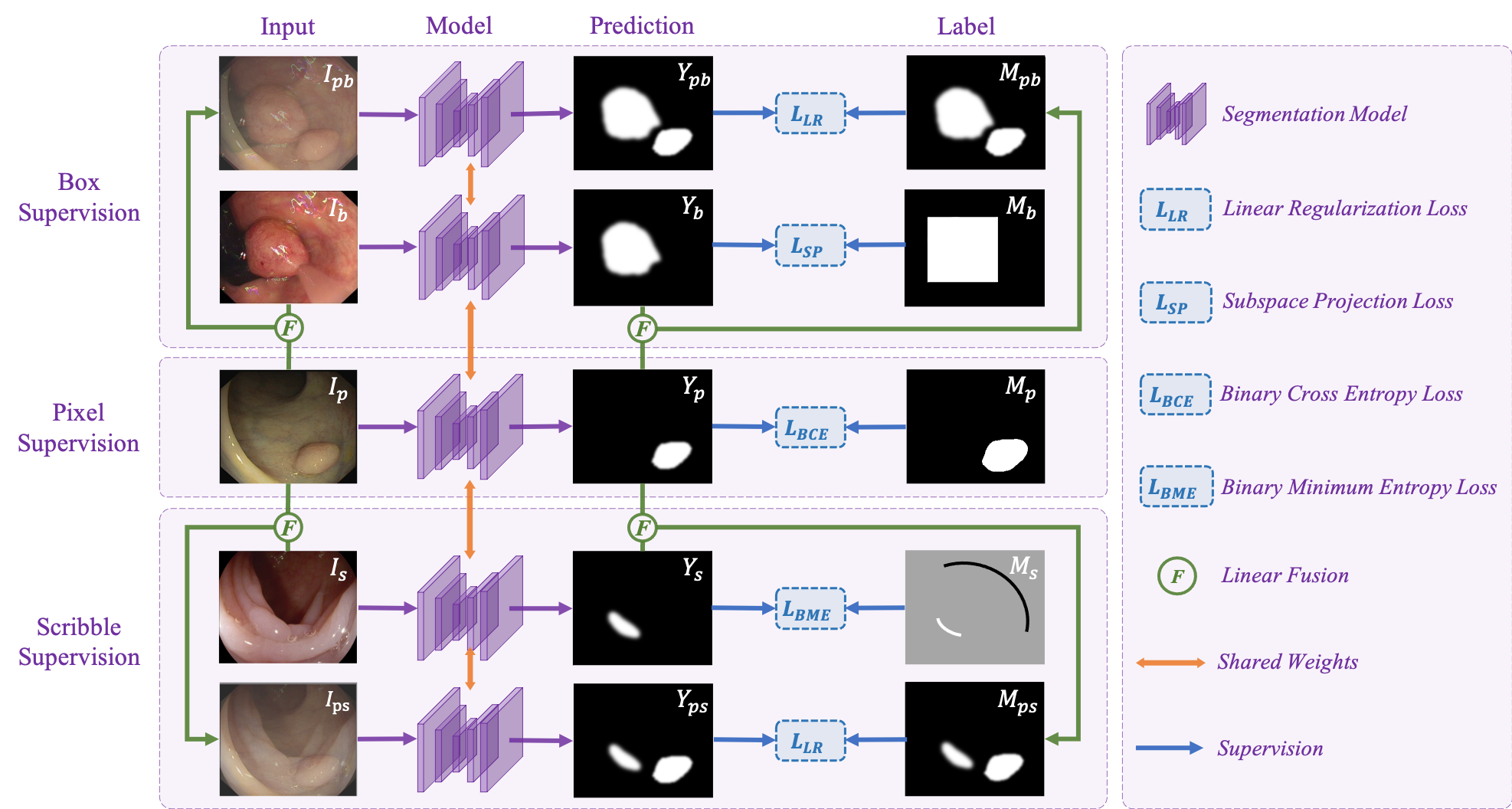}
\caption{\textbf{Illustration of our MixPolyp framework.} It consists of three learning branches: (1) Full supervision branch for pixel-level annotation data, (2) Box supervision branch with Subspace Projection loss ($\mathcal{L_{SP}}$), and (3) Scribble supervision branch with Binary Minimum Entropy loss ($\mathcal{L_{BME}}$). In addition, in both box supervision and scribble supervision branches, we introduce Linear Regularization loss ($\mathcal{L_{LR}}$) to constrain the consistency between predictions.
}
\label{fig:pipeline}
\end{figure*}

Colorectal cancer is a prevalent cancer worldwide, posing a serious threat to human health. Fortunately, automated polyp segmentation methods have been developed in recent years. For example, U-Net~\cite{unet} has achieved significant performance through pixel-wise supervision. However, these models face challenges such as data scarcity and overfitting due to the high cost of acquiring annotations. To address these limitations, weakly supervised methods have been explored, leveraging less precise annotations (bounding boxes and scribbles). While these approaches reduce the annotation burden, they often fail to fully exploit the wealth of available data.

In this context, we propose \textbf{MixPolyp}, a mixed supervision polyp segmentation model (Fig.~\ref{fig:pipeline}) designed to overcome the limitations of existing methods. Unlike fully or weakly supervised models that rely on a single type of label, MixPolyp integrates pixel-, box-, and scribble-level annotations, thereby expanding the available data. This combination of supervision types not only reduces annotation costs but also improves model generalization. Furthermore, box and scribble annotations are less prone to noise caused by the ambiguous boundaries of polyps, mitigating the impact of subjective labeling errors and making them a more efficient and practical alternative for large-scale clinical applications.

While training segmentation models using various annotations is promising, integrating these annotations presents significant challenges. Box annotations misclassify some background pixels as polyp ones, leading to performance degradation. Scribble annotations are sparse and provide insufficient supervision, leaving most pixels unlabeled. In response to this, we propose the novel MixPolyp model, selectively leveraging the strengths of box and scribble annotations while mitigating their drawbacks. MixPolyp incorporates three key components: Subspace Projection loss ($\mathcal{L_{SP}}$), Binary Minimum Entropy loss ($\mathcal{L_{BME}}$), and Linear Regularization loss ($\mathcal{L_{LR}}$).

$\mathcal{L_{SP}}$ corrects shape inconsistencies between the prediction and box annotations by projecting them into 1D vectors along horizontal and vertical axes and then calculating the supervision loss to reduce shape discrepancies.
$\mathcal{L_{BME}}$ tackles the sparse supervision of scribble annotations, particularly in unlabeled regions where traditional binary cross-entropy loss is ineffective. It computes the loss for all possible labels of unlabeled pixels and selects the minimum value as the supervisory loss. 
Given the limitations of $\mathcal{L_{SP}}$ and $\mathcal{L_{BME}}$ in producing unique predictions, we introduce $\mathcal{L_{LR}}$ to provide dense supervision. It blends fully and weakly annotated images to create synthetic images, ensuring that their predictions align with the combined outputs of both annotation types.

In summary, MixPolyp redesigns the supervision loss for weakly annotated data without altering the model structure, making it a versatile approach that can be integrated with other models. Besides, these losses are only adopted during training, incurring no computational cost during inference. Although very simple, MixPolyp surprisingly predicts high-quality polyp masks, outperforming previous fully supervised results. Our contributions are three-fold: (1) We propose a mixed supervised polyp segmentation paradigm that fully leverages diverse annotation types, greatly expanding data availability while reducing labeling costs; (2) We design the subspace projection loss, binary minimum entropy loss, and linear regularization loss to tackle annotation noise and supervision sparsity; (3) We conduct extensive experiments on five datasets, demonstrating the effectiveness of our approach.

\section{Method}
\label{method}
Fig.~\ref{fig:pipeline} illustrates the pipeline of MixPolyp. For the input image $I\in R^{H \times W}$, the backbone network extracts four scales of features $\{f_i | i=1,...,4 \}$ with the resolutions $[\frac{H}{2^{i+1}}, \frac{W}{2^{i+1}}]$. To balance accuracy and efficiency, only $f_2$, $f_3$, and $f_4$ are used. These features are unified in channel dimensions via $1 \times 1$ convolutions, resized with bilinear upsampling, and multiplied together before a final $1 \times 1$ convolution for prediction. To reduce the reliance on precise annotations, MixPolyp integrates three types of annotated data: a small set of images $I_p$ with pixel-level annotations $M_p$, a large set of images $I_b$ with box-level annotations $M_b$, and another large set of images $I_s$ with scribble-level annotations $M_s$. During training, triplets $(I_p, I_b, I_s) \in \mathbb{R}^{H \times W \times 3}$ are fed into MixPolyp, yielding corresponding predictions $(Y_p, Y_b, Y_s) \in \mathbb{R}^{H \times W}$.

\subsection{Box Supervision with Subspace Projection Loss}
\label{sec:box}
\begin{figure}[t]
    \centering
    \includegraphics[width=\linewidth]{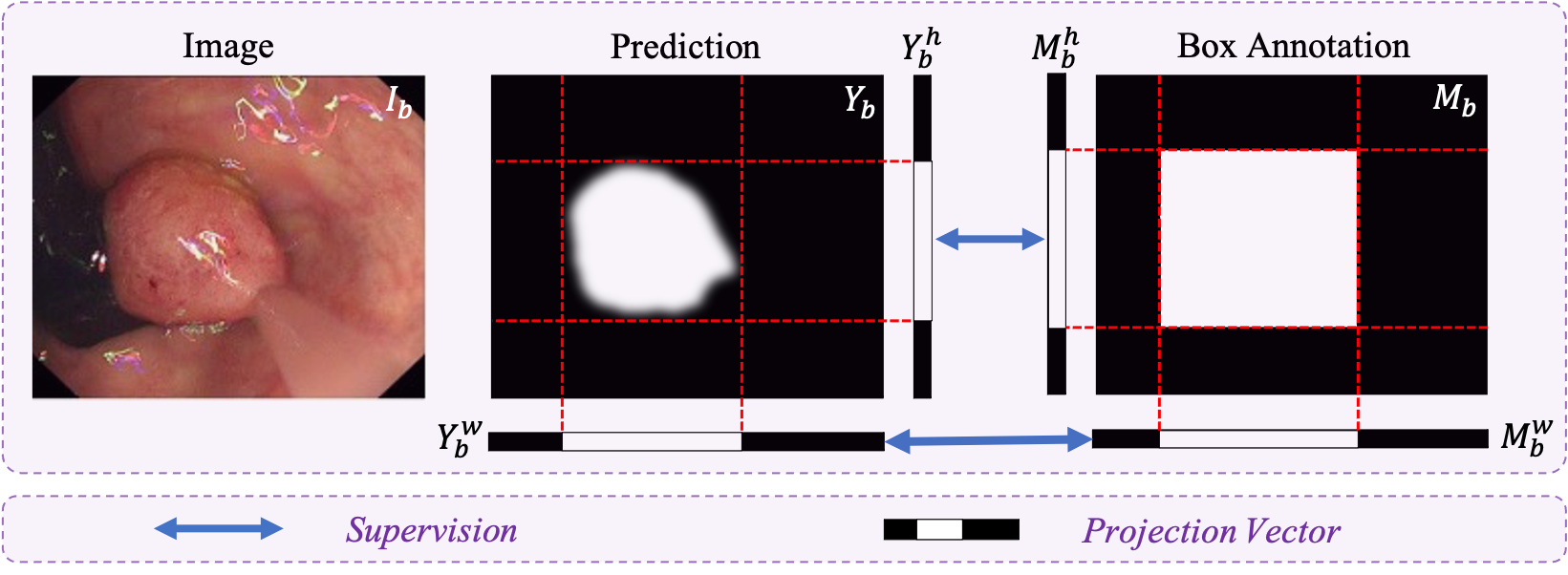}
    \caption{Subspace Projection Loss, which first projects the predicted mask and the box annotations into 1D vectors and then calculates the supervision loss between these vectors.}
    \label{fig:loss:lsp}
\end{figure}

\begin{table*}[t]
    \centering
    \caption{Performance comparison with different polyp segmentation models. The \textcolor{red}{red} column represents the weighted average (wAVG) performance of different testing datasets. Next to the dataset name is the image quantity of each dataset.}
    \label{tab:performance}
    \renewcommand\tabcolsep{7.5pt}
    \begin{tabular}{lcccccccccccc}
    \toprule
        & \multicolumn{2}{c}{\textbf{ColonDB} (380)} & \multicolumn{2}{c}{\textbf{Kvasir} (100)} & \multicolumn{2}{c}{\textbf{ClinicDB} (62)} & \multicolumn{2}{c}{\textbf{EndoScene} (60)} & \multicolumn{2}{c}{\textbf{ETIS} (196)} & \multicolumn{2}{c}{\textcolor{red}{\textbf{wAVG} (798)}} \\
        \cmidrule(l){2-3}\cmidrule(l){4-5}\cmidrule(l){6-7}\cmidrule(l){8-9}\cmidrule(l){10-11}\cmidrule(l){12-13}
    \multirow{-3}{*}{\textbf{Methods}}   & Dice & IoU  & Dice & IoU  & Dice & IoU  & Dice & IoU  & Dice & IoU  & \textcolor{red}{Dice} & \textcolor{red}{IoU} \\ 
    \midrule
    U-Net~\cite{unet}             & 51.2\% & 44.4\% & 81.8\% & 74.6\% & 82.3\% & 75.0\% & 71.0\% & 62.7\% & 39.8\% & 33.5\% & \textcolor{red}{56.1\%} & \textcolor{red}{49.3\%}\\
    PraNet~\cite{pranet}          & 70.9\% & 64.0\% & 89.8\% & 84.0\% & 89.9\% & 84.9\% & 87.1\% & 79.7\% & 62.8\% & 56.7\% & \textcolor{red}{74.0\%} & \textcolor{red}{67.5\%}\\
    MSNet~\cite{zhao2021automatic}        & 75.1\% & 67.1\% & 90.5\% & 84.9\% & 91.8\% & 86.9\% & 86.5\% & 79.9\% & 72.3\% & 65.2\% & \textcolor{red}{78.5\%} & \textcolor{red}{71.4\%}\\
    SANet~\cite{sanet}            & 75.3\% & 67.0\% & 90.4\% & 84.7\% & 91.6\% & 85.9\% & 88.8\% & 81.5\% & 75.0\% & 65.4\% & \textcolor{red}{79.4\%} & \textcolor{red}{71.4\%}\\  
    Polyp-Pvt~\cite{polyppvt}     & 80.8\% & 72.7\% & 91.7\% & 86.4\% & 93.7\% & 88.9\% & 90.0\% & 83.3\% & 78.7\% & 70.6\% & \textcolor{red}{83.3\%} & \textcolor{red}{76.0\%}\\
    LDNet~\cite{ldnet}            & 79.4\% & 71.5\% & 91.2\% & 85.5\% & 92.3\% & 87.2\% & 89.3\% & 82.6\% & 77.8\% & 70.7\% & \textcolor{red}{82.2\%} & \textcolor{red}{75.1\%}\\
    SSFormer-S~\cite{ssformer}    & 77.2\% & 69.7\% & 92.5\% & 87.8\% & 91.6\% & 87.3\% & 88.7\% & 82.1\% & 76.7\% & 69.8\% & \textcolor{red}{81.0\%} & \textcolor{red}{74.3\%}\\
    SSFormer-L~\cite{ssformer}    & 80.2\% & 72.1\% & 91.7\% & 86.4\% & 90.6\% & 85.5\% & 89.5\% & 82.7\% & 79.6\% & 72.0\% & \textcolor{red}{83.0\%} & \textcolor{red}{75.7\%}\\
    HSNet~\cite{zhang2022hsnet}   & 81.0\% & 73.5\% & 92.6\% & 87.7\% & 94.8\% & 90.5\% & 90.3\% & 83.9\% & 80.8\% & 73.4\% & \textcolor{red}{84.2\%} & \textcolor{red}{77.4\%}\\ 
    \rowcolor{black!10}
    \textbf{MixPolyp (Ours)}             & 82.8\% & 75.0\% & 92.3\% & 87.1\% & 92.5\% & 87.6\% & 90.5\% & 83.5\% & 85.0\% & 76.4\% & \textbf{\textcolor{red}{85.9\%}} & \textbf{\textcolor{red}{78.5\%}}\\
    \bottomrule
\end{tabular}
\end{table*}

For data with box annotations, a naive approach is to convert bounding boxes into masks $M_b$ to supervise predicted masks $Y_b$. However, this method often results in poor generalization due to the shape bias in $M_b$. To overcome this limitation, we propose the indirect supervision loss $\mathcal{L_{SP}}$, which avoids the misleading of rectangular shape bias in the annotations by transforming $Y_b$ and $M_b$ into a shape-independent space for supervision (Fig.~\ref{fig:loss:lsp}). The implementation of $\mathcal{L_{SP}}$ is as follows:

\textbf{Projection.} The predicted mask $Y_b$ is projected horizontally and vertically into two vectors, $Y_b^w \in [0,1]^{1 \times W}$ and $Y_b^h \in [0,1]^{H \times 1}$, using max pooling to capture lesion position and extent. Box annotations $M_b$ are similarly projected into $M_b^w$ and $M_b^h$. This process preserves key lesion features while removing shape bias.

\textbf{Supervision.} By projecting $(Y_b^w, Y_b^h)$ and $(M_b^w, M_b^h)$ into 1D vectors, $\mathcal{L}_{SP}$ resolves shape inconsistencies and mitigates noise in box annotations. As shown in Eq.~\ref{eq:loss:BD}, we compute supervision using binary cross-entropy loss $\mathcal{L_{BCE}}=-\sum_{i,j}[M_{i,j}log(Y_{i,j})+(1-M_{i,j})log(1-Y_{i,j})]$ and Dice loss $\mathcal{L_{DICE}}=1-2\frac{\sum_{i,j}{M_{i,j}Y_{i,j}}}{\sum_{i,j}(M_{i,j}+Y_{i,j})}$. $\mathcal{L}_{SP}$ is fully differentiable and can be seamlessly integrated into models for efficient gradient back-propagation using PyTorch.
\begin{equation}
  \begin{aligned}
    \mathcal{L}_{SP}  = & 0.5[\mathcal{L_{BCE}}(Y_b^w, M_b^w)+\mathcal{L_{BCE}}(Y_b^h, M_b^h))] + \\
    & 0.5[\mathcal{L_{DICE}}(Y_b^w, M_b^w)+\mathcal{L_{DICE}}(Y_b^h, M_b^h)].
    \label{eq:loss:BD}
  \end{aligned}
\end{equation}

\subsection{Scribble Supervision with Binary Minimum Entropy Loss}
\label{sec:scribble}
\begin{figure}[t]
    \centering
    \includegraphics[width=\linewidth]{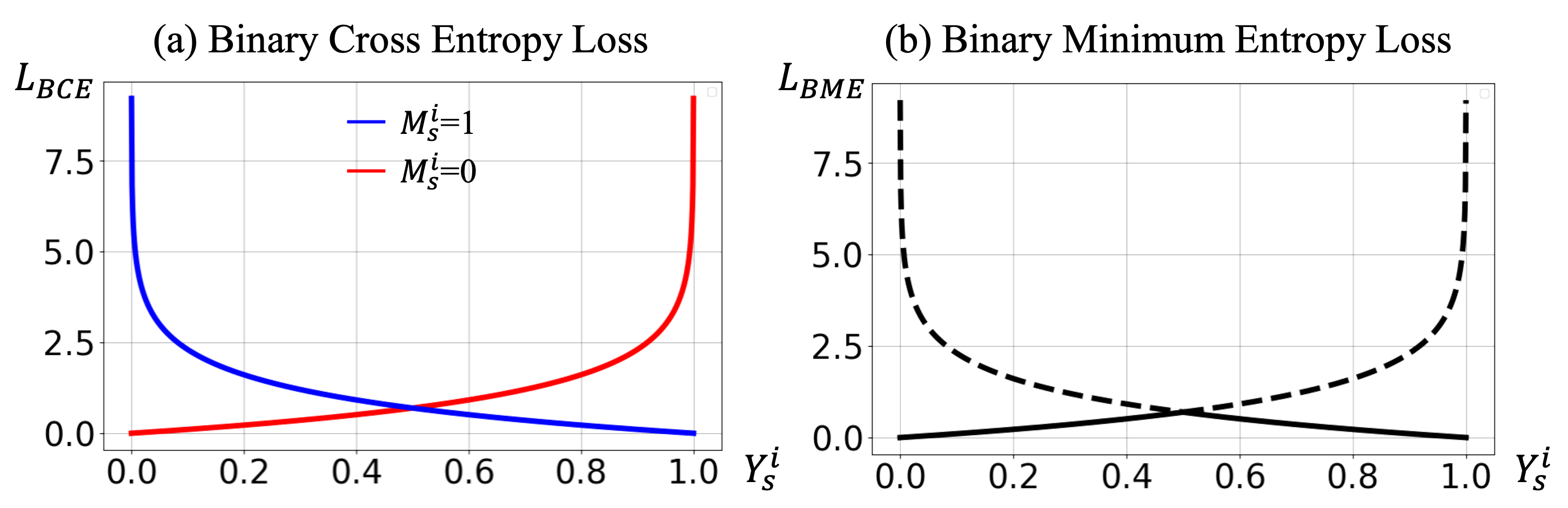}
    \caption{Loss curve comparison between Binary Cross Entropy Loss and Binary Minimum Entropy Loss.}
    \label{fig:loss:bme}
\end{figure}

In scribble-level supervision, prior methods apply $\mathcal{L_{BCE}}$ to only a small subset of labeled pixels, leading to non-unique predictions. To provide supervision for the majority of unlabeled pixels, we propose the Binary Minimum Entropy loss ($\mathcal{L_{BME}}$), which computes $\mathcal{L_{BCE}}$ for all possible labels ({\it i.e.}, $M_s^i=0$ and $M_s^i=1$) and selects the minimum as its loss value, as shown in Eq.~\ref{equ:bme}, where $Y_s^i$ and $M_s^i$ denote the prediction and label at location $i$, and $U$ represents the subset of pixels outside scribbles.
\begin{align}
    &\mathcal{L_{BME}}(Y_s^i)_{i \in U}   = \min(-\log(Y_s^i), -\log(1-Y_s^i)).
    \label{equ:bme}
\end{align}

$\mathcal{L_{BME}}$ is based on the minimum entropy principle, encouraging the model to make confident predictions by reducing entropy in the unlabeled regions. Fig.~\ref{fig:loss:bme}(b) illustrates the loss curve (solid line) of $\mathcal{L_{BME}}$. As predictions approach 0 or 1, the loss decreases, promoting confidence. Unlike $\mathcal{L_{BCE}}$ requiring explicit labels, $\mathcal{L_{BME}}$ offers supervision without labels. The sparse supervision $\mathcal{L_{BCE}}$ from scribble pixels guide the model towards accurate predictions, and dense supervision $\mathcal{L_{BME}}$ from unlabeled pixels enhance the model's accuracy and robustness. This combined loss $\mathcal{L}_{scribble}$ (Eq.~\ref{equ:loss:scribble}) ensures that the model benefits from all pixels in the input image.
\begin{align}
    \mathcal{L}_{scribble} = \frac{\sum\limits_{i\in U}\mathcal{L_{BME}}(Y_s^i, M_s^i)+\sum\limits_{j\in S}\mathcal{L_{BCE}}(Y_s^j, M_s^j)}{|U|+|S|},
    \label{equ:loss:scribble}
\end{align}
where $\mathcal{S}$ denotes the subset of scribble pixels.

\subsection{Linear Regularization Loss}
\label{sec:mix}
During training, the fully supervised branch benefits from finely annotated data, allowing it to quickly learn accurate target features. However, it is prone to overfitting due to the lack of data. Conversely, the weakly supervised branch struggles to converge due to sparse annotations. To enhance model robustness, we propose two hybrid supervision branches, where predictions from the fully supervised branch are used to generate pseudo-labels. We introduce Linear Regularization loss ($\mathcal{L}_{LR}$) to assist the weakly supervised branch, ensuring both branches complement each other for improved generalization. The linear regularization loss is defined as:
\begin{align}
    & Y_{ph} = \mathcal{F}(Y_p, Y_h) = (\lambda \cdot Y_p + (1-\lambda)\cdot Y_h),\\
    & \mathcal{L}_{LR} = -\frac{1}{\mathcal{D}}\sum_{i \in \mathcal{D}}|Y_{ph}-M_{ph}|, ~ h \in b,s,
\end{align}
where $Y_p$ and $Y_h$ are the predictions from the fully and weakly supervised branches ($h=b$ for box, $h=s$ for scribble supervision). $Y_{ph}$ and $M_{ph}$ are the predictions and pseudo-labels in the hybrid branch. $\mathcal{F}$ denotes the linear fusion, $\lambda$ controls the fusion, and $D$ is the set of pixels in the image.

As shown in Fig.~\ref{fig:pipeline}, image pairs containing fully and weakly supervised data are combined via linear fusion $\mathcal{F}$ with a weight $\lambda$ to generate simulated data $I_{pb}$ or $I_{ps}$. Then, we simultaneously train the network with both simulated and real data, producing the prediction $Y_{ph}$ at every iteration. Subsequently, we apply the same fusion strategy to the predictions of real data to generate pseudo-labels $M_{ph}$, which transfer fine-grained features from the fully supervised branch to the weakly supervised branch, compensating for sparse annotations and improving overall model performance.

\subsection{Overall Training Loss Function}
\label{sec:loss}
For the fully supervised branch, we use the combination loss $\mathcal{L}_{pixel}=\mathcal{L_{BCE}} + \mathcal{L_{DICE}}$. The total loss $\mathcal{L}_{total}$ would be:
\begin{align}
    \mathcal{L}_{total} = \mathcal{L}_{pixel} + \mathcal{L}_{SP}+\mathcal{L}_{scribble} + \mathcal{L}_{LR}.
    \label{loss:total}
\end{align}

\section{Experiment}
\subsection{Dataset and Implementation Details}
In this study, we utilize seven datasets: Kvasir~\cite{kvasir}, CVC-ClinicDB~\cite{clinicdb}, CVC-ColonDB~\cite{colondb}, EndoScene~\cite{endoscence}, ETIS~\cite{etis}, SUN-SEG~\cite{sunseg}, and LDPolypVideo~\cite{ldpolypvideo}. Following PraNet~\cite{pranet}, the first five datasets, comprising 1,451 training images, are used with pixel-level annotations for full supervision. LDPolypVideo, containing 33,884 samples, provides box annotations, while SUN-SEG, with 49,136 samples, supplies scribble annotations. Here, We employ PVTv2-B2~\cite{wang2021pvtv2} as the backbone and implement the model using PyTorch. All the training images are uniformly resized to $352 \times 352$ and perform random flip and random rotation as data augmentation following SANet~\cite{sanet}. The model is optimized with SGD, using a momentum of 0.9, an initial learning rate of 0.05, and a batch size of 16, over 50,000 iterations.

\begin{table}[t]
    \centering
    \caption{The ablation studies of MixPolyp with various loss functions.}
    \renewcommand\tabcolsep{10pt}
    \begin{tabular}{cccc|cc}
        \toprule
        $\mathcal{L_{BCE}}$& $\mathcal{L_{SP}}$ & $\mathcal{L_{BME}}$ & $\mathcal{L_{LR}}$ & Dice & IoU \\
        \hline
        $\checkmark$ & & & & 80.6\% & 72.6\% \\
        $\checkmark$ & $\checkmark$ & & & 84.0\% & 76.6\% \\
        $\checkmark$ & & $\checkmark$ & & 84.3\% & 76.8\% \\
        $\checkmark$ & $\checkmark$ & $\checkmark$ & & 84.9\% & 77.4\% \\
        \rowcolor{black!10}
        $\checkmark$ & $\checkmark$ & $\checkmark$ & $\checkmark$ & \textbf{85.9\%} & \textbf{78.5\%} \\
        \bottomrule
    \end{tabular}
    \label{tab:ablation}
\end{table}
\subsection{Performance Comparison}
As shown in Table~\ref{tab:performance}, we compare MixPolyp against 9 state-of-the-art models across 5 datasets. MixPolyp consistently outperforms competing methods, achieving the highest weighted average Dice (85.9\%) and IoU (78.5\%). This represents a notable improvement over the second-best model, HSNet, surpassing it by 1.7\% on Dice and 1.1\% on IoU. The ablation study in Table~\ref{tab:ablation} assesses the impact of the proposed loss functions on MixPolyp’s performance. Each row shows the Dice and IoU scores for different loss combinations. The baseline model with only $\mathcal{L_{BCE}}$ achieves a Dice score of 80.6\% and an IoU of 72.6\%. Adding the Subspace Projection loss ($\mathcal{L_{SP}}$) improves the performance to a Dice of 84.0\% and IoU of 76.6\%, while the Binary Minimum Entropy loss ($\mathcal{L_{BME}}$) further increases these to 84.3\% and 76.8\%. Incorporating the Linear Regularization loss ($\mathcal{L_{LR}}$) results in a Dice of 84.9\% and IoU of 77.4\%. The full model, using all the loss functions, achieves the best performance with a Dice of 85.9\% and IoU of 78.5\%, demonstrating the effectiveness of each loss and their combined benefits in improving segmentation accuracy.


\section{Conclusion}
MixPolyp addresses data scarcity in polyp segmentation by combining mask, box, and scribble annotations, significantly reducing labeling costs and increasing available data. The proposed modules are implemented only during training, ensuring no added inference overhead. Experiments on multiple datasets demonstrate MixPolyp's superior performance. Future work will explore incorporating additional annotation types to enhance the model's capabilities.

\section{Acknowledgement}
The work was in part supported by NSFC (Tianyuan Fund for Mathematics) with Grant No. 12326610 and the Shenzhen Science and Technology Program with Grant No. JCYJ20220818100015031.

\bibliographystyle{IEEEtran}
\bibliography{IEEEabrv,bibtex}

\begin{thebibliography}{10}
\providecommand{\url}[1]{#1}
\csname url@samestyle\endcsname
\providecommand{\newblock}{\relax}
\providecommand{\bibinfo}[2]{#2}
\providecommand{\BIBentrySTDinterwordspacing}{\spaceskip=0pt\relax}
\providecommand{\BIBentryALTinterwordstretchfactor}{4}
\providecommand{\BIBentryALTinterwordspacing}{\spaceskip=\fontdimen2\font plus
\BIBentryALTinterwordstretchfactor\fontdimen3\font minus \fontdimen4\font\relax}
\providecommand{\BIBforeignlanguage}[2]{{%
\expandafter\ifx\csname l@#1\endcsname\relax
\typeout{** WARNING: IEEEtran.bst: No hyphenation pattern has been}%
\typeout{** loaded for the language `#1'. Using the pattern for}%
\typeout{** the default language instead.}%
\else
\language=\csname l@#1\endcsname
\fi
#2}}
\providecommand{\BIBdecl}{\relax}
\BIBdecl

\bibitem{unet}
O.~Ronneberger, P.~Fischer, and T.~Brox, ``U-net: Convolutional networks for biomedical image segmentation,'' in \emph{MICCAI}, 2015, pp. 234--241.

\bibitem{pranet}
D.-P. Fan, G.-P. Ji, T.~Zhou, G.~Chen, H.~Fu, J.~Shen, and L.~Shao, ``Pranet: Parallel reverse attention network for polyp segmentation,'' in \emph{MICCAI}, 2020, pp. 263--273.

\bibitem{zhao2021automatic}
X.~Zhao, L.~Zhang, and H.~Lu, ``Automatic polyp segmentation via multi-scale subtraction network,'' in \emph{MICCAI}, 2021.

\bibitem{sanet}
J.~Wei, Y.~Hu, R.~Zhang, Z.~Li, S.~K. Zhou, and S.~Cui, ``Shallow attention network for polyp segmentation,'' in \emph{MICCAI}, 2021.

\bibitem{polyppvt}
B.~Dong, W.~Wang, D.-P. Fan, J.~Li, H.~Fu, and L.~Shao, ``Polyp-pvt: Polyp segmentation with pyramid vision transformers,'' \emph{arXiv preprint arXiv:2108.06932}, 2021.

\bibitem{ldnet}
R.~Zhang, P.~Lai, X.~Wan, D.-J. Fan, F.~Gao, X.-J. Wu, and G.~Li, ``Lesion-aware dynamic kernel for polyp segmentation,'' in \emph{MICCAI}, 2022, pp. 99--109.

\bibitem{ssformer}
J.~Wang, Q.~Huang, F.~Tang, J.~Meng, J.~Su, and S.~Song, ``Stepwise feature fusion: Local guides global,'' in \emph{MICCAI}, 2022, pp. 110--120.

\bibitem{zhang2022hsnet}
W.~Zhang, C.~Fu, Y.~Zheng, F.~Zhang, Y.~Zhao, and C.-W. Sham, ``Hsnet: A hybrid semantic network for polyp segmentation,'' \emph{Computers in biology and medicine}, vol. 150, p. 106173, 2022.

\bibitem{kvasir}
D.~Jha, P.~H. Smedsrud, M.~A. Riegler, P.~Halvorsen, T.~de~Lange, D.~Johansen, and H.~D. Johansen, ``Kvasir-seg: A segmented polyp dataset,'' in \emph{MultiMedia modeling}, 2020, pp. 451--462.

\bibitem{clinicdb}
J.~Bernal, F.~J. S{\'a}nchez, G.~Fern{\'a}ndez-Esparrach, D.~Gil, C.~Rodr{\'\i}guez, and F.~Vilari{\~n}o, ``Wm-dova maps for accurate polyp highlighting in colonoscopy: Validation vs. saliency maps from physicians,'' \emph{Computerized Medical Imaging and Graphics}, vol.~43, pp. 99--111, 2015.

\bibitem{colondb}
J.~Bernal, J.~S{\'a}nchez, and F.~Vilarino, ``Towards automatic polyp detection with a polyp appearance model,'' \emph{Pattern Recognition}, vol.~45, no.~9, pp. 3166--3182, 2012.

\bibitem{endoscence}
D.~V{\'a}zquez, J.~Bernal, F.~J. S{\'a}nchez, G.~Fern{\'a}ndez-Esparrach, A.~M. L{\'o}pez, A.~Romero, M.~Drozdzal, and A.~Courville, ``A benchmark for endoluminal scene segmentation of colonoscopy images,'' \emph{Journal of healthcare engineering}, vol. 2017, 2017.

\bibitem{etis}
J.~Silva, A.~Histace, O.~Romain, X.~Dray, and B.~Granado, ``Toward embedded detection of polyps in wce images for early diagnosis of colorectal cancer,'' \emph{IJCSRS}, vol.~9, no.~2, pp. 283--293, 2014.

\bibitem{sunseg}
G.-P. Ji, G.~Xiao, Y.-C. Chou, D.-P. Fan, K.~Zhao, G.~Chen, and L.~Van~Gool, ``Video polyp segmentation: A deep learning perspective,'' \emph{Machine Intelligence Research}, pp. 1--19, 2022.

\bibitem{ldpolypvideo}
Y.~Ma, X.~Chen, K.~Cheng, Y.~Li, and B.~Sun, ``Ldpolypvideo benchmark: A large-scale colonoscopy video dataset of diverse polyps,'' in \emph{MICCAI}, 2021, pp. 387--396.

\bibitem{wang2021pvtv2}
W.~Wang, E.~Xie, X.~Li, D.-P. Fan, K.~Song, D.~Liang, T.~Lu, P.~Luo, and L.~Shao, ``Pvtv2: Improved baselines with pyramid vision transformer,'' \emph{CVMJ}, vol.~8, no.~3, pp. 1--10, 2022.

\end{thebibliography}

\end{document}